\documentclass[journal,twoside]{IEEEtran}
\ifCLASSINFOpdf
\else
\fi

\usepackage{graphicx}
\usepackage{amsmath}
\usepackage{amssymb}
\usepackage[colorlinks]{hyperref}
\usepackage{verbatim}

\def\0{{\bf 0}}
\def\1{{\bf 1}}

\def\etal{{\em et al.}}
\def\eg{{\em e.g.}}
\def\ie{{\em i.e.}}

\def\etal{{\em et al.\/}\,}

\graphicspath{{Figs/}}

\usepackage{tabularx} 
\usepackage{booktabs}   
\usepackage{mathrsfs} 
\usepackage{multirow} 
\usepackage{amsfonts} 
\usepackage{wrapfig}  
\usepackage[misc]{ifsym}
\usepackage{cite}
\usepackage{subcaption}

\begin{document}

\title{Attention-Guided Autoencoder for Automated Progression Prediction of Subjective Cognitive Decline with Structural MRI}

\author{Hao~Guan, \and
Ling~Yue, \and
Pew-Thian~Yap, \and
Shifu~Xiao, 
Andrea Bozoki,
\and
        Mingxia~Liu,~\IEEEmembership{Senior Member,~IEEE}
\thanks{H.~Guan, P.-T.~Yap and M.~Liu are with the Department of Radiology and Biomedical Research Imaging Center, University of North Carolina at Chapel Hill, Chapel Hill, NC 27599, USA. 
L.~Yue and S.~Xiao are with the Department of Geriatric Psychiatry, Shanghai Mental Health Center, Shanghai Jiao Tong University School of Medicine, Shanghai 200030, China. 
A.~Bozoki is with the Department of Neurology, University of North Carolina at Chapel Hill, NC 27599, USA. 
}
\thanks{Corresponding authors: M.~Liu (mxliu@med.unc.edu) and L.~Yue (bellinthemoon@hotmail.com).}

}

\if false
\markboth{IEEE TRANSACTIONS ON BIOMEDICAL ENGINEERING,~Vol.~XX, No.~XX, October~2022}%
{H. Guan,  and M. Liu: Early Identification of Progressive Disease through Transfer Learning}
\fi

\maketitle

\begin{abstract}
Subjective cognitive decline (SCD) is the preclinical stage of Alzheimer's disease (AD) which happens even earlier than mild cognitive impairment (MCI). Progressive SCD will convert to MCI with the potential of further evolving to AD. Therefore, early identification of progressive SCD with neuroimaging techniques (\eg, structural MRI) is of great clinical value for early intervention of AD. 
However, existing MRI-based machine/deep learning methods usually suffer the small-sample-size problem and lack interpretability.
To this end, we propose an interpretable autoencoder model with domain transfer learning (IADT) for progression prediction of SCD.
Firstly, the proposed model can leverage MRIs from both the target domain (\ie, SCD) and auxiliary domains (\eg, AD and NC) for progressive SCD identification.
Besides, it can automatically locate the disease-related brain regions of interest (defined in brain atlases) through an attention mechanism, which shows good interpretability.
In addition, the IADT model is 
straightforward to train and test with only 5$\thicksim$10 seconds on CPUs and is suitable for medical tasks with small datasets.
Extensive experiments on the publicly available ADNI dataset and a private CLAS dataset have demonstrated the effectiveness of the proposed method.

\end{abstract}


%
\IEEEpeerreviewmaketitle


\section{Introduction}

\begin{figure}[!tbp]
\setlength{\belowcaptionskip}{-2pt}
\setlength{\abovecaptionskip}{1pt}
\setlength{\abovedisplayskip}{-2pt}
\setlength{\belowdisplayskip}{-2pt}
\center
 \includegraphics[width= 1.0\linewidth]{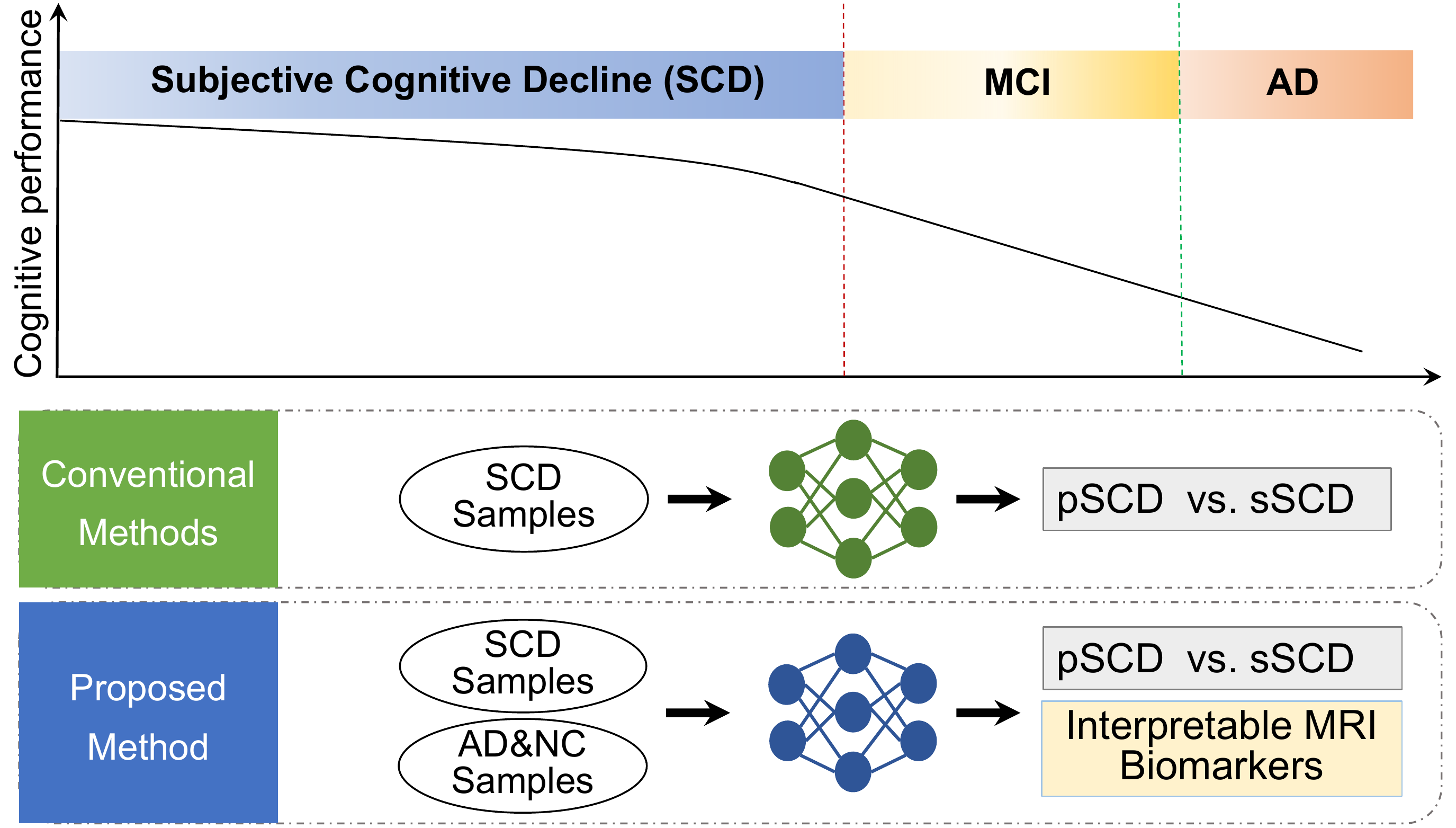}
 \caption{Progression of AD pathology~\cite{jessen2014conceptual} (top) and the differences of our method from conventional methods (bottom). MCI: mild cognitive impairment; SCD: subjective cognitive decline; p/sSCD: progressive SCD; sSCD: stable SCD.} 
 \label{fig_SCD}
\end{figure}

\IEEEPARstart{A}s one of the main causes of dementia and death, Alzheimer's disease (AD) has affected millions of older people around the world~\cite{AD}.
As illustrated in Fig.~\ref{fig_SCD}, AD is characterized as a chronic neurodegenerative process with an extended spectrum. Its prodromal stage is mild cognitive impairment (MCI), while an even earlier preclinical stage is termed as subjective cognitive decline (SCD) or subjective memory complaint (SMC)\cite{jessen2014conceptual,rabin2017subjective}.
Growing evidence has verified that individuals with SCD may suffer higher risks of evolving to AD.
Thus progression prediction of SCD is of great clinical value for early intervention of the brain disease progress.
Neuroimaging has been shown as an effective technology for understanding the mechanisms of different brain disorders and has already been adopted for the progression prediction of SCD~\cite{wang2020neuroimaging,parker2022systematic}.

Structural magnetic resonance imaging (sMRI) is a widely used imaging modality for AD-spectrum research~\cite{ADNI,CADDementia}.
As the preclinical stage of AD, it typically takes several years for a progressive SCD (pSCD) subject to evolve to MCI, which makes the data collection a time-consuming and challenging task. 
Although several previous studies have applied machine learning to MRI-based SCD progression prediction~\cite{yue2021prediction,felpete2020predicting,lin2022identification,liu2022assessing}, they usually suffer from the small-sample-size problem. 
Compared with small-sized SCD samples, there are much more AD and normal control (NC) samples, such as those in the public ADNI database~\cite{ADNI}. 
Thus, it is interesting to employ these relatively large-scale AD and NC samples to assist the task of SCD progression prediction. 
From a clinical point of view, as shown in Fig.~\ref{fig_SCD}, AD typically goes through the following stages (extended spectrum):
NC$\rightarrow$stable SCD (sSCD)$\rightarrow$ pSCD$\rightarrow$stable MCI (sMCI)$\rightarrow$progressive MCI (pMCI)$\rightarrow$AD. 
Across the full spectrum of AD, sSCD is close to NC while pSCD is close to AD, thus the model trained on AD and NC samples could be helpful to assist the task of pSCD vs. sSCD classification. 
Several studies~\cite{zhang2022single,guan2021learning,lian2018hierarchical} have revealed that machine/deep learning models trained on AD and NC samples can achieve good results when applied directly to the task of pMCI vs. sMCI classification. 
In this work, we propose to leverage AD and NC samples for the training for pSCD vs. sSCD classification. 
In addition, we are interested in finding important brain regions-of-interest (ROIs) that are associated with SCD progression, thus helping enhance the interpretability of learning-based diagnostic systems.

\begin{figure*}[!tbp]
\setlength{\belowcaptionskip}{1pt}
\setlength{\abovecaptionskip}{1pt}
\center
 \includegraphics[width= 0.98\linewidth]{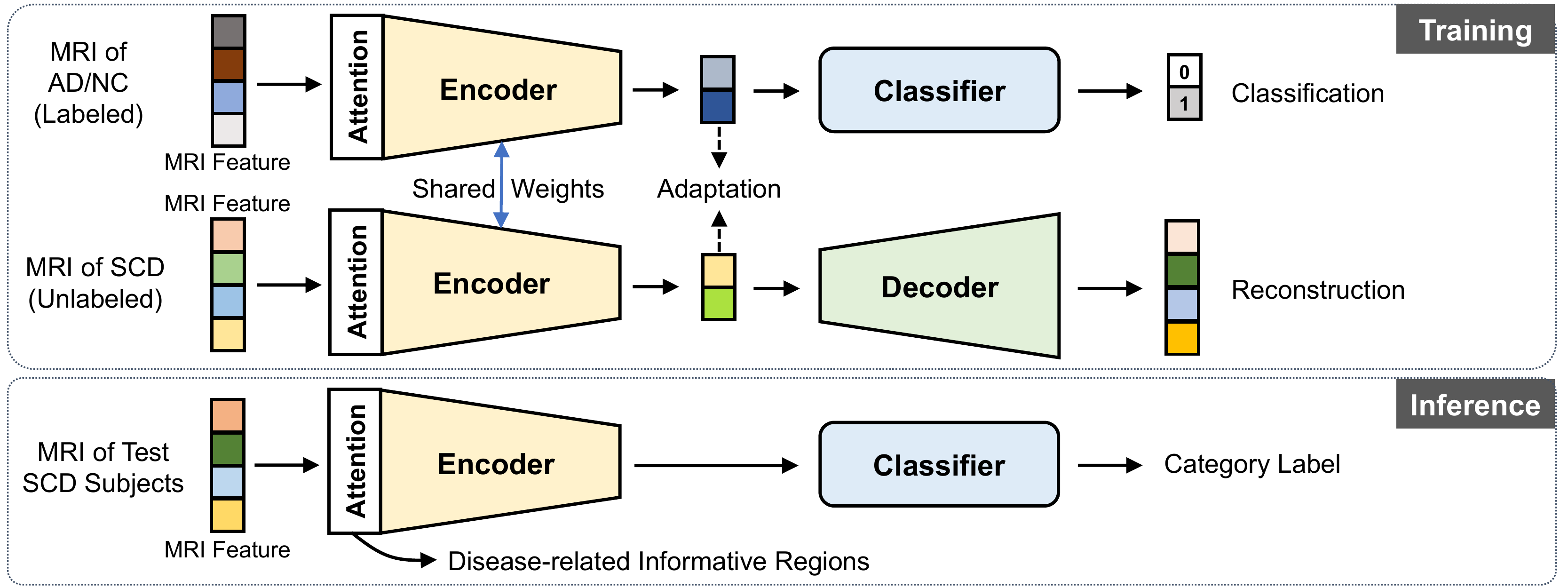}
 \caption{Illustration of the main framework of the proposed IADT model for MRI-based SCD progression prediction.}
 \label{fig_Main}
\end{figure*}

Based on these motivations, we propose an interpretable autoencoder model with domain transfer learning (IADT) with application to SCD progression prediction. 
As illustrated in Fig.~\ref{fig_Main}, the IADT model includes four key components: (1) a feature encoder that learns the shared subspace representations of different domains, (2) an attention component that helps detect disease-related brain regions of interest, (3) a decoding module that reconstructs the original input, 
(4) a classification module that is responsible for the identification of brain disorders.
During training, the labeled AD/NC samples are used to train the classification module, while unlabeled SCD samples are used to train the reconstruction module. 
A domain adaptation loss is utilized to enforce the distribution of source domain (AD and NC) and target domain (pSCD and sSCD) to get close. Through training with these losses, the network can learn some features that are both discriminative to brain disorders and reflect the properties of SCD, thus can be applied to pSCD vs. sSCD classification.

The contributions of this work are listed as follows:
\begin{itemize}
\vspace{2pt}
\item This paper proposes an interpretable autoencoder model with attention mechanism which can automatically locate the important brain regions that are related to subjective cognitive decline. This helps enhance interpretability of MRI biomarkers learned by deep learning to improve their utility in clinical practice. 

\item The proposed network structure can leverage MRIs from both the target domain (\ie, SCD) and auxiliary source domains (\eg, AD and NC) for SCD progression prediction to increase their utility in clinical practice and improve clinical diagnosis. 

\item To evaluate the effectiveness of the proposed IADT model, we conduct experiments on the ADNI and CLAS datasets. Our method achieves state-of-the-art results on SCD progression prediction with good interpretability. 

\end{itemize}

The rest part of this paper is structured as follows.
Related studies are first reviewed in Section \ref{Related_works}. The materials used in this work are introduced in Section~\ref{Materials}.
Section~\ref{Methodology} introduces the proposed method in details. 
Section~\ref{Experiment} presents experimental settings, evaluation metrics, competing methods, and experimental results. The brain ROIs highlighted by the model and the influence of several key parameters of the proposed method are analyzed in Section~\ref{Discussion}. 
The paper is finally concluded in Section \ref{Conclusion}.

\if false
This work makes the following three contributions.
\begin{itemize}
  \item \textbf{Dataset}. We build a ...
\end{itemize}
\fi

\section{Related Works} \label{Related_works}
\subsection{Learning-based SCD Progression Prediction} 
As a preclinical stage of AD, SCD has attracted growing awareness in the field
of brain disease analysis.
Existing studies on SCD can be roughly divided into non-neuroimaging-based methods and neuroimaging-based methods. 
Engedal~\etal\cite{engedal2020power} explore electroencephalography (EEG) with a statistical method to predict SCD conversion.
Engedal~\etal\cite{pereiro2021relevance} use demographic information through machine learning to predict the progression of SCD.
Since structural MRIs have been widely used in brain disorder identification, numerous machine learning methods have been proposed to deal with AD-related brain disorder classification~\cite{mirzaei2016imaging,tanveer2020machine}. 
Yue~\etal\cite{yue2021prediction} utilize cost-sensitive support vector machine (CSVM) for SCD progression prediction based on clinical and MRI features.
Felpete~\etal\cite{felpete2020predicting} use SVM and random forest for the prediction of SCD conversion.
Lin~\etal\cite{lin2022identification} introduce sparse coding for MRI feature selection and use random forest for SCD progression prediction.
Liu~\etal\cite{liu2022assessing} propose a GAN-based framework to synthesize PET and MRI data of SCD which can increase the training data and assist SCD progression prediction.
Despite the progress, identifying progressive SCD (pSCD) and stable SCD (sSCD) individuals is still a challenging task due to the small-sample-size problem and insignificant pathological brain changes.
Since SCD is a preclinical stage of AD, how to leverage the knowledge from AD/NC classification (with a relatively large amount of labeled samples, and significant pathological brain changes) to SCD progression prediction is an open problem with clinical value.

\subsection{Domain Adaptation for AD-related Disease Identification}
Transfer learning~\cite{cheplygina2019not,valverde2021transfer,TL_1} and domain adaptation~\cite{guan2022domain,csurka2017domain} have been used to tackle problems when the target domain has relatively fewer data and different distribution from the source domain.
These methods do not require identical distribution of training/source data and test/target data and can share knowledge between different domains.
Some studies use transfer learning for early diagnosis of AD.
Cheng~\etal\cite{cheng2015domain} propose a transferable support vector machine with cross-domain kernel learning for MCI conversion prediction. 
With the recent progress of deep learning (DL), some methods leverage deep neural network through fine-tuning to facilitate transfer learning for AD analysis~\cite{mehmood2021transfer}.
Lian~\etal\cite{lian2018hierarchical} reveal that training a deep network with AD and NC samples is beneficial for MCI conversion prediction.
These transfer learning methods indicate that AD has inherent relationship with its early stages (\eg, MCI), thus we explore leveraging knowledge learned from AD and NC for predicting SCD progression.

\section{Materials} \label{Materials}
\subsection{Data Acquisition}
We employ T1-weighted structural MRIs from two datasets in this work for model training and test.
The first one is the publicly available brain MRI dataset, \ie, Alzheimer's Disease Neuroimaging Initiative (ADNI)~\cite{ADNI}\footnote{https://ida.loni.usc.edu}.
We use 3T T1-weighted structural MRIs of 159 AD patients and 201 normal controls (NC) from the ADNI dataset in this work. 
The second dataset is the Chinese Longitudinal Aging Study (CLAS) dataset~\cite{CLAS}. This dataset includes 76 SCD subjects with 3T T1-weighted structural MRIs.
There are 24 progressive SCD (pSCD) and 52 stable SCD (sSCD) subjects. 
These pSCD subjects have evolved to MCI within the next 84 months, while the sSCD ones remain stable.  
The demographic information for these two datasets is shown in Table~\ref{tab:demographic}. 
To analyze the statistical difference, we use hypothesis testing at the 5\% significance level, while the null hypothesis is ``there is no significant difference".
Specifically, we adopt paired \textit{t}-test to evaluate the difference of these two groups (\ie, pSCD and sSCD) in terms of age. 
The \textit{p}-value is 0.1177, which accepts the null hypothesis at the 5\% significance level.
This indicates that statistically the ages of these two cohorts, \ie, pSCD and sSCD,  have no significant difference.

We also analyze the relationship between the classification output and the age, cognition scores (MMSE scores).
Specifically, the age and MMSE are set as the input variables, and classification result (pSCD: 1, sSCD: 0) is the response output. We adopt logistic regression to fit the relationship between the input (age, MMSE) and output (pSCD or sSCD). We got the {\em p}-values: 0.2405 and 0.1578 for the age and MMSE, respectively, which indicates that the coefficients for age and MMSE are not significant for pSCD/sSCD.

\begin{table}[t]
\setlength{\belowcaptionskip}{1pt}
\setlength{\abovecaptionskip}{1pt}
\caption{Demographic information of the subjects for SCD
progression prediction. The values 
are denoted in the form of ``mean$\pm$standard deviation''.
F/M: Female/Male.}
\centering
\setlength{\tabcolsep}{3mm}{
\begin{tabular} {lllll} 
\toprule[1.2pt]
Dataset                  &Category   &Gender (F/M)   &Age       &MMSE\\

\midrule

\multirow{2}*{ADNI}     &AD        &67/92         &74.4$\pm$8.1    &23.1$\pm$2.2 \\
                        &NC        &105/95        &73.4$\pm$6.2    &29.0$\pm$1.3 \\  
                        &pSMC      &9/12          &73.9$\pm$5.9    &27.8$\pm$2.0\\
                        &sSMC      &55/35         &71.9$\pm$5.4    &28.5$\pm$1.9\\
\midrule

\multirow{2}*{CLAS}    &pSCD      &13/11    &71.3$\pm$6.6    &26.8$\pm$2.6\\
                       &sSCD      &27/25    &68.6$\pm$7.2    &27.7$\pm$2.2\\

\bottomrule[1.2pt]
\end{tabular}
}
\label{tab:demographic}
\end{table}

\subsection{Image Preprocessing and MRI Feature Extraction}
All the T1-weighted MRIs are preprocessed through a standard pipeline,
including 1) skull stripping, 2)  intensity inhomogeneity correction, 3) registration to the Automated Anatomical Labeling (AAL) template~\cite{AAL}, and 4) re-sampling to the resolution of $1 \times 1 \times 1$ mm$^3$.
We use Freesurfer\footnote{https://surfer.nmr.mgh.harvard.edu/} to facilitate skull stripping and adopt the SPM software package\footnote{https://www.fil.ion.ucl.ac.uk/spm/} to conduct the other MRI preprocessing steps. All the processed MRIs have the identical dimension of $181\times217\times181$.

For statistical analysis and model training, we extract the ROI features from 
MRIs. 
We adopt the AAL brain atlas\footnote{https://www.gin.cnrs.fr/en/tools/aal/} for feature extraction.
The AAL atlas is an anatomical parcellation of the spatially normalized T1 volume which has labeled $90$ brain regions of interest. 
During computation, the AAL atlas is used as a mask.
The features of a brain MRI are then calculated as the volumes of gray matter in each of the $90$ brain regions. 
The dimension of the ROI feature is $90$, and each of the dimension reflects the gray matter volume in the corresponding brain region.
The advantage of ROI features is that they contain prior knowledge from neuroscientists and have good interpretability for indicating which brain areas are more closely related to the disease classification. 
\section{Methodology} \label{Methodology}
\subsection{Problem Formulation}
We leverage domain adaptation for SCD progression prediction. 
Suppose the joint feature space of samples and the corresponding labels are represented as $\mathcal{X} \times \mathcal{Y}$. We define a source domain ${\mathcal{D}_S}$ and a target domain ${\mathcal{D}_T}$ on the feature space, but they have different distributions. A total of $n_s$ samples with category labels are provided in the source domain, \ie, $\mathcal{D}_S = \{(\mathbf{x}^S_i, y^S_i)\}^{n_s}_{i=1}$, while a number of $n_t$ samples are given the target domain \ie, $\mathcal{D}_T = \{(\mathbf{x}^T_j)\}^{n_t}_{j=1}$. 
The same set of category labels are shared by the source and target domains. 
In our work, AD and NC samples are set as the source domain (AD: 1, NC: 0), while pSCD and sSCD samples are the target domain (pSCD: 1, sSCD: 0).
The goal is to train a model with labeled source domain samples and unlabeled/labeled target domain samples so the learned model can generalize well to target data for label estimation.

\subsection{Interpretable Autoencoder with Domain Transfer}
As illustrated in Figure~\ref{fig_Main}, the proposed interpretable autoencoder with domain transfer (IADT) framework consists of 4 modules. 
(1) The attention component automatically detects the most discriminative brain MRI features (ROIs) of the input data. 
(2) The encoder projects the input data from the original feature space to a latent compressed feature space. 
(3) The decoder reconstructs the input data from the intermediate compressed representations to its original feature space. 
(4) The classifier predicts the category labels. 
In the training phase, the source data and target data first go through the attention module for ``feature filtering''. Then the data are fed into the encoder module and projected into a low-dimensional latent feature space. A maximum mean discrepancy (MMD) loss is facilitated on these compressed representations of source and target data to reduce their domain difference.
With the compressed representations, a classifier is trained with the category labels (source domain) to enforce the projected latent features more discriminative to brain disorders.
Meanwhile, a decoder is trained using reconstruction loss to enforce the projected latent features containing the most useful information within the target data. 
In the test stage, the trained classifier is directly applied to the target domain (SCD) for prediction 
(\ie, pSCD vs. sSCD classification).

\subsection{Attention Mechanism for Brain ROI Selection}
\begin{figure}[!tbp]
\setlength{\belowcaptionskip}{1pt}
\setlength{\abovecaptionskip}{1pt}
\center
 \includegraphics[width= 0.96\linewidth]{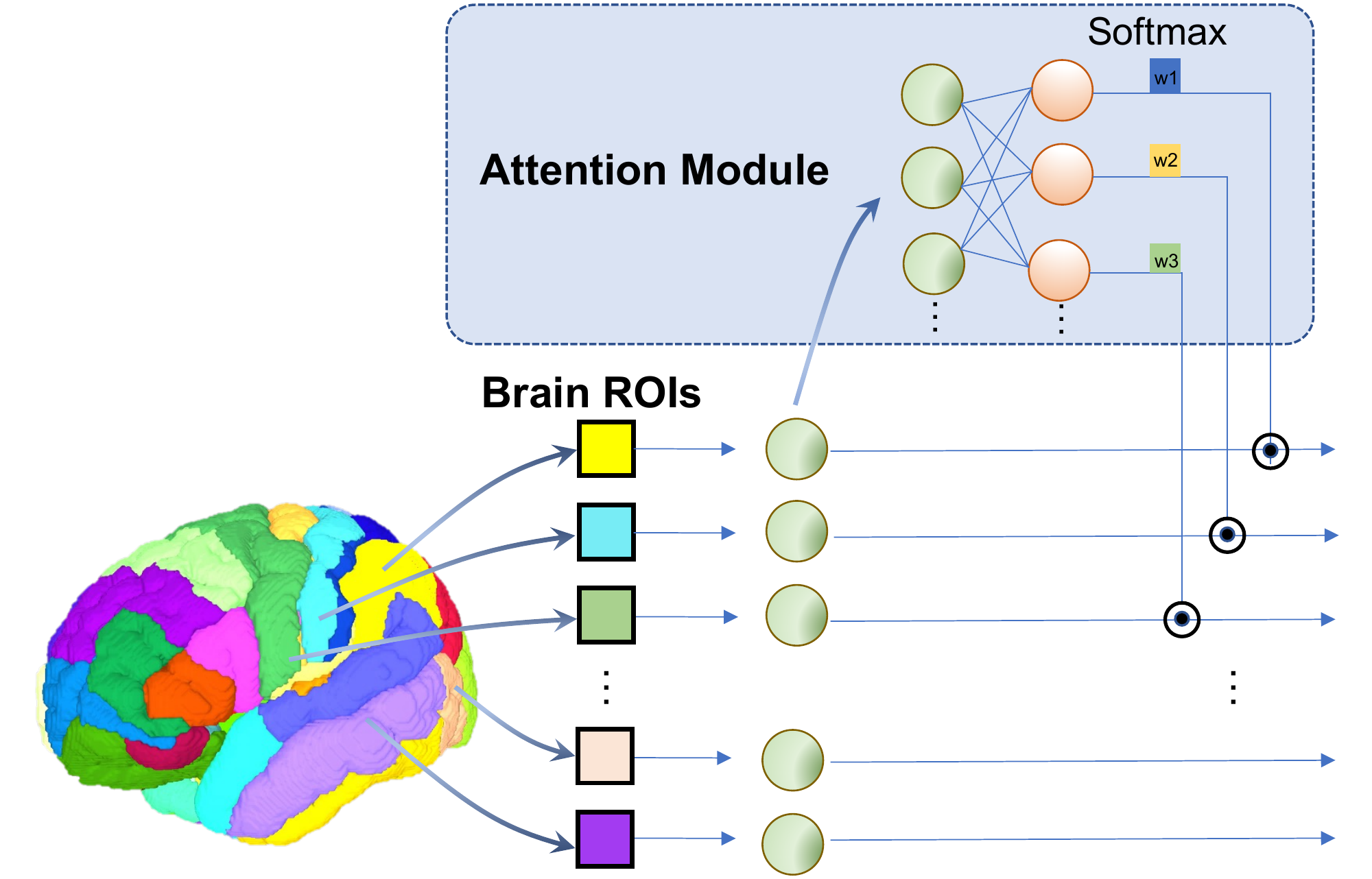}
 \caption{Illustration of the attention module of the proposed IADT model for SCD progression prediction.}
 \label{fig_Attention}
\end{figure}

Brain disorders have a close relationship with specific brain areas~\cite{pegueroles2017,wenk2003,mu2011adult,ott2010brain}.
Since we use brain ROI feature as input, we aim to identify which regions are more disease-related automatically to enhance  interpretability of MRI biomarkers. 
As illustrated in Fig.~\ref{fig_Attention}, the 90-dimensional ROI feature is used as the network input, denoted as $I = ({R_1, R_2, \cdots, R_n})$. The feature is fed into an attention network that consists 90 neurons. 
The softmax is used as the activation function to
output a probability (weight) on each dimension of the 90-dimension feature with a summation of 1, \ie, $\mathbf{w} = [w_1, w_2, \cdots, w_n], \sum_{i=1}^{n} w_i = 1$. 
These weights are then multiplied with the original input ROI feature to get weighted features. From the output probabilities of the attention module, we can analyze the importance of different brain regions.

\subsection{Feature Encoding and Adaptation}
After feature importance weighting by the attention module, data from the source and target domains are fed into the feature encoder. The feature encoder projects the data into a low-dimensional latent space. To achieve this, we design a three-layer multiple-layer-perception (MLP) with 90, 64, 32 neurons, respectively. 
After encoding, all the data have been projected to the 32-dimensional space. The encoders for the source and target domains share the same weights.
To align the feature distributions of source and target data in the projected space, we adopt maximum mean discrepancy (MMD) loss which is defined as:
\begin{equation}
\mathbf{MMD}^2_k = \left \| \mathbf{E}_p [\phi (\mathbf{x}^s)  ] -  \mathbf{E}_q [\phi (\mathbf{x}^t)  ]               \right\|^2_{\mathcal{H}_k}
\label{mmd_loss}
\end{equation}
where ${\mathcal{H}_k}$ represents the Reproducing Kernel Hilbert Space (RKHS) endowed with a kernel function $k$, and
$k(\mathbf{x}^s, \mathbf{x}^t) =\left \langle \phi (\mathbf{x}^s), \phi (\mathbf{x}^s) \right \rangle$.
By minimizing the MMD loss, the source and target data distributions in the latent space can get closer.

\subsection{Joint Classification and Reconstruction}
In our model, the encoder module plays the role of learning domain-sharing features. To this end, the training of the encoder should be driven by tasks with certain supervisions. 
For AD/NC samples, due to their relatively large amount and abundant label information, a classification module is trained with these labeled source data with the cross-entropy loss as:
\begin{equation}
\mathcal{L}_{cls} = \sum\nolimits_{i=1}^N -y_i ln (\hat{y}_i) - (1-y_i) ln (1- \hat{y}_i)
\label{cls_loss}
\end{equation}
where $y_i$ is the label (0, 1) and $\hat{y}_i$ is the output of the classifier. 
For target data, \ie, SCD samples, since there is no label information provided during training, we use the reconstruction loss as supervision. Specifically, a decoder is linked to the output of the encoder, trying to reconstruct the original input target MRI. The decoder has a symmetric structure to the encoder, with 32, 64, 90 neurons in each layer, respectively. We use $\mathcal{L}_1$ norm in the reconstruction loss: 
\begin{equation}
\mathcal{L}_{recon} = \sum\nolimits_{i=1}^N \left \| \mathbf{x}^t_i - \mathbf{\hat{x}}^t_i   \right\|_1
\label{recon_loss}
\end{equation}
where  $\mathbf{x}^t_i$ denotes the $i$-th sample in the target domain while $\mathbf{\hat{x}}^t_i $ is the corresponding reconstructed result by the decoder.

The classification and reconstruction modules are trained jointly and they provide supervision for the learning process. Through the joint training, the encoder is encouraged to learn features that are both discriminative to brain disorders and reflect properties of SCD data.
It should be noted that the total number of target samples (SCD) is much smaller than the source data (AD/NC). 
To facilitate joint training, the SCD samples must be enhanced. Here we simply duplicate the SCD samples to make them have an equal number with the AD/NC samples. This can make the classification and reconstruction modules updated each training epoch. 
To avoid a bias towards the AD/NC data, the reconstruction loss for SCD data can be allocated a relatively higher weight during training. 
The overall loss $\mathcal{L}$ for training is calculated as:
\begin{equation}
\mathcal{L} = \lambda_1 \mathcal{L}_{mmd} +  \lambda_2 \mathcal{L}_{cls} +  \mathcal{L}_{recon}
\label{overall_loss}
\end{equation}
where $\lambda_1$ and $\lambda_2$ are the parameters to control the contributions of the three terms in Eq.~\eqref{overall_loss}. 

\subsection{Implementation}
The proposed IADT is implemented by the PyTorch software package.
The Adam optimizer is utilized for training with a learning rate of $0.001$. 
A linear kernel is adopted as the kernel function for the MMD loss in Eq.~\eqref{mmd_loss}.
The parameters $\lambda_1$ and $\lambda_2$ in Eq.~\eqref{overall_loss} are set to 0.1 and 0.1, respectively. 
The model is trained for 60 epochs with the batch size of $128$.
The overall training and test take around 5$\thicksim$10 seconds.

\begin{table}[t]
\setlength{\belowcaptionskip}{1pt}
\setlength{\abovecaptionskip}{1pt}
\caption{Performance (\%) of eight methods for SCD progression prediction (\ie, pSCD vs. sSCD classification), and $p$-values via paired sample $t$-test between our method and each of the competing methods.}
\centering
{
\begin{tabular} {llllll|c} 
\toprule[1.2pt]
Method      &ACC     &BAC    &AUC   &SEN   &SPE  &$p<0.05$\\
\midrule
Baseline    &57.89       &54.65      &58.17     &45.83     &63.46   &Yes\\
TCA         &60.53       &59.94      &55.85     &58.33     &61.54   &Yes\\   
GFK         &59.21       &57.85      &58.81     &54.17     &61.54   &Yes\\                         
SA          &61.84       &59.78      &60.26     &54.17     &65.38   &Yes\\
CORAL       &59.21       &60.10     &56.01    &62.50    &57.69  &Yes\\
TL          &64.47       &59.46     &60.50    &45.83    &73.08  &Yes\\
VoxCNN      &67.11       &54.65     &55.45      &20.83      &\textbf{88.46} &Yes\\
IADT (Ours)        &\textbf{71.05} &\textbf{69.87}    &\textbf{64.90}  &\textbf{66.67} &73.08 &--\\
\bottomrule[1.2pt]
\end{tabular}
}
\label{tab:result}
\end{table}

\section{Experiment}  \label{Experiment}
\subsection{Experimental Setup}
In our experiment, we adopt the ADNI dataset as the source domain, while the CLAS dataset as the target domain. 
The model architecture keeps fixed throughout the experiment. 

For performance evaluation, we utilize five metrics, \ie, classification accuracy (ACC), balanced accuracy (BAC), sensitivity (SEN), specificity (SPE), and area under the ROC curve (AUC). 
Let ${\rm TP}$, ${\rm TN}$, ${\rm FP}$, ${\rm FN}$ represent true positive, true negative, false positive and false negative, respectively. 
Then each metric is calculated as follows. ${\rm ACC}= \frac{{\rm TP+TN}}{{\rm TP+TN+FP+FN}}$, ${\rm SEN} = \frac{{\rm TP}}{{\rm TP+FN}}$, ${\rm SPE} = \frac{{\rm TN}}{{\rm TN+FP}}$, and ${\rm BAC} = \frac{{\rm SEN+SPE}}{{\rm 2}}$. 
Please note that a higher value for each metric indicates a better classification performance.

\subsection{Competing Methods}
The proposed method is compared with the following seven methods, including one baseline method, four statistical learning methods, and two deep learning methods.

(1) \textbf{Baseline}.
This method uses logistic regression, one of the most popular classification models in neuroimaging analysis~\cite{logistic1,logistic2,wachinger2016domain} for structural MRI-based SCD progression prediction.
It is trained on the AD/NC samples from ADNI, and then applied to CLAS for pSCD/sSCD classification. 

(2) \textbf{Transfer component analysis (TCA)}~\cite{TCA}. 
In TCA, source and target data are projected to the subspace spanned by several transfer components.
These transfer components are learned in the reproducing kernel Hilbert space using maximum mean discrepancy (MMD). 
After that, a logistic classifier trained with source data (AD/NC) is used to facilitate pSCD/sSCD classification in the projected feature space. 
We use a linear kernel for feature learning. 
The subspace dimension of TCA is set to $40$, with $\lambda=0.01$ and $\gamma=0.1$. 

(3) \textbf{Geodesic flow kernel (GFK)}~\cite{GFK}.
In this method, a domain specific n-dimensional subspace is calculated for the source and target data. Then the source and target data are projected into an intermediate subspace along the shortest geodesic path connecting these two n-dimensional subspaces on the Grassmann manifold.
After adaptation, a logistic classifier is adopted to conduct pSCD vs. sSCD classification using these projected data. 
In our experiment, we set the subspace dimension in GFK to $20$.

(4) \textbf{Subspace alignment (SA)}~\cite{SA}.
In this method, the source data is projected into the source subspace and the target data is represented by the target subspace. Both of them use their principal components as the subspace representation.
Then a transformation matrix that maps the source subspace to the target one is learned to mitigate domain shift.
After adaptation, a logistic classifier is trained to do pSCD vs. sSCD classification using these projected data. 
The subspace dimension in SA is set to $20$.

(5) \textbf{Correlation alignment (CORAL)}~\cite{CORAL}.
In this method, domain shift is reduced through alignment of the second-order statistics of source and target domains. 
After adaptation, a logistic classifier is used for pSCD vs. sSCD classification using these aligned data. 
CORAL only calculates the covariance of source and target features without any additional parameters.

(6) \textbf{Normal transfer learning (TL)}. 
In this method, a four-layer multiple-layer-perception (MLP) with 90, 64, 32, and 2 neurons is used as the network architecture of the learning model. The network is firstly pretrained with AD and NC samples, and then fine-tuned with 10\% of the pSCD and sSCD samples in the CLAS dataset. The fine-tuned network is then applied to the remaining pSCD and sSCD samples for testing.

(7) \textbf{VoxCNN~\cite{VoxCNN}}. VoxCNN is a deep 3D convolutional neural network for structural MRI-based brain disorder classification. It is composed of $10$ convolution layers for feature learning. The size of filter kernel is $3\times3\times3$, and the number of filters for each layer is 8, 8, 16, 16, 32, 32, 32, 64, 64, and 64, respectively. Two fully-connected layers are utilized for classification.
Note that this network takes 3D volumetric data (MRIs) as input.
We first train the VoxCNN with AD and NC samples from the ADNI dataset, then directly apply it for pSCD/sSCD classification on the CLAS dataset.

\subsection{Results of SCD Progression Prediction}
We conduct experiment on the ADNI and CLAS datasets for SCD progression prediction, \ie, pSCD vs. sSCD classification. 
Label information of CLAS is not available during our model training. 
The performance of different methods is listed in Table~\ref{tab:result}. From Table~\ref{tab:result}, we have the following observations.

\emph{First}, in terms of ACC and BAC, the baseline has an inferior performance to the other competing methods. This indicates that domain shifts among different domains can significantly affect a machine learning model. Meanwhile, the adaptation-based methods can achieve better results than the baseline, which implies the effectiveness of domain adaptation. 
\emph{Second}, the proposed IADT model outperforms the other methods by a large margin. This can be attributed to two reasons. 1) Most competing methods conduct adaptation through alignment of the source and target in an unsupervised way, \ie, the label information of the source data is not utilized in the adaptation process. We argue that incorporating the label information of source domain enables the model to be more discriminative. 2) The attention mechanism in our model can select more important features for classification which is helpful for training.

To further evaluate the generalization ability of the proposed method, we apply the well-trained model to unseen SMC samples from ADNI, without any model retraining.
T1-weighted MRIs of 21 progressive SMC (pSMC) and 90 stable SMC (sSMC) samples from ADNI are used. 
All these MRIs are pre-processed in the same way as those from CLAS.
We conduct pSMC vs. sSMC classification using our model and compare its performance with the baseline (\ie, logistic classifier) and report the results in Fig.~\ref{fig_SCD_ADNI}.  
As can be seen from this figure, the proposed method has good generalization ability on independent ADNI data.

\begin{table*}[!tbp]
\setlength{\belowcaptionskip}{1pt}
\setlength{\abovecaptionskip}{1pt}
\caption{Performance comparison of the proposed model and state-of-the-art methods for SCD progression prediction.}
\centering
\setlength{\tabcolsep}{4mm}{
\begin{tabular} {lllllll|c} 
\toprule[1.2pt]
Method    &Model   &ACC (\%)    &BAC (\%)   &AUC (\%)   &SEN (\%)   &SPE (\%)  &$p<0.05$\\
\midrule
Yue~\etal\cite{yue2021prediction}     &Cost-Sensitive SVM         &53.95     &56.25 &51.52       &62.50      &50.00   &Yes\\
Felpete~\etal\cite{felpete2020predicting}     &Random Forest      &59.21     &57.85 &63.66          &54.17    &61.54   &Yes\\
Liu~\etal\cite{liu2020joint}   &GAN   &65.50         &67.05        &\textbf{71.30}          &\textbf{72.50}      &61.60   &--\\
IADT (Ours)    &Attention Autoencoder    &\textbf{71.05}    &\textbf{69.87}         &64.90          &66.67        &\textbf{73.08}  &--\\
\bottomrule[1.2pt]
\end{tabular}
}
\label{tab:result2}
\end{table*}

\begin{figure}[!th]
\setlength{\belowcaptionskip}{1pt}
\setlength{\abovecaptionskip}{1pt}
\center
 \includegraphics[width= 0.7\linewidth]{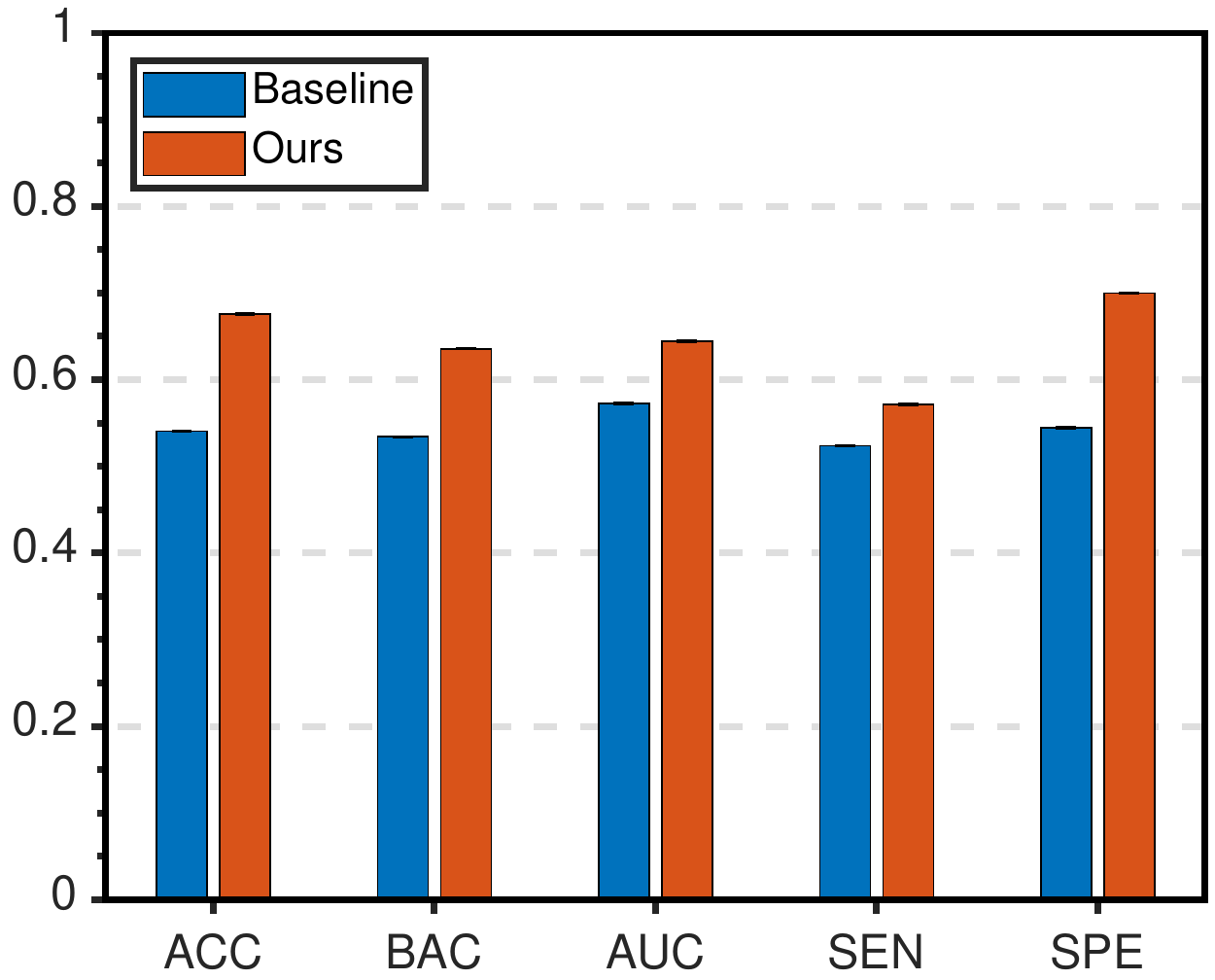}
 \vspace{4pt}
 \caption{Results of pSMC vs. sSMC classification on ADNI.} 
 \label{fig_SCD_ADNI}
\end{figure}

\subsection{Comparison with State-of-the-Art Methods}

Our method is further compared with several state-of-the-arts for structural MRI-based pSCD vs. sSCD classification. These methods include: 1) Cost-Sensitive SVM (CSVM)~\cite{yue2021prediction}, 2) Random Forest~\cite{felpete2020predicting} and 3) Generative Adversarial Network~\cite{liu2020joint}. 
We reproduce these methods 
and test them on the same dataset.
As for the CSVM, it is implemented through an SVM with a linear kernel, and is trained with a cost matrix $\big(\begin{smallmatrix}
  0 & 10\\
  1 & 0
\end{smallmatrix}\big)$. 
The RF is built through an ensemble of decision tree models. The GAN is composed of a generator network with 3 convolution layers, 3 residual blocks and 2 deconvolution layers, and a discriminator network with 5 convolution layers. 

The prediction results are reported in Table~\ref{tab:result2}. From Table~\ref{tab:result2}, our method achieves better or comparable performance than the state-of-the-art methods. More specifically, our model achieves
higher accuracy, balanced accuracy (\ie, 71.05\%, 69.87\%) and specificity (\ie, 73.08\%) which are much better than the other three state-of-the-art methods, even though GAN is a complex deep-learning method. 
Note that the GAN model utilizes both MRI and PET data with much more training data (both original and synthetic images) than ours. Thus it achieves a higher SEN and AUC value. 
Despite this, our model still achieves comparable performance. Considering that the proposed model only takes 5$\thicksim$10 seconds on CPUs to train while the GAN typically takes several days, our model has made a good balance between accuracy and efficiency.
\subsection{Results on MCI Conversion Prediction}
We evaluate the proposed model in the task of MCI to AD conversion prediction, \ie, pMCI vs. sMCI classification.
A total of 393 MCI subjects (167 pMCI, 226 sMCI) with structural MRIs are used as the unlabeled target domain, while AD and NC samples are adopted as the labeled source domain. 
The pMCI subjects have converted to AD within 36 months while sMCI have not. 
We still extract ROI features to represent each subject.
The VoxCNN directly takes MRIs as the input for end-to-end training. It is firstly pre-trained with AD and NC samples and then applied for MCI conversion prediction.
The results of different methods for MCI conversion prediction are reported in Table~\ref{tab:MCI}.
From the results, we have the following observations.
1) Our method achieves better or comparable results than the other methods. This verifies the effectiveness of our method for early identification of Alzheimer's disease. 
2) The VoxCNN achieves a good result. This is consistent with related studies~\cite{lian2018hierarchical,guan2021learning}, implying a deep CNN directly trained with MR images is able to perform well on MCI tasks. Despite that, the proposed model still achieves better results and makes a balance between effectiveness and efficiency.

\begin{table}[t]
\setlength{\belowcaptionskip}{1pt}
\setlength{\abovecaptionskip}{1pt}
\caption{Performance (\%) of seven methods for MCI-to-AD conversion prediction (\ie, pMCI vs. sMCI classification).}
\centering
{
\begin{tabular} {llllll|c} 
\toprule[1.2pt]
Method      &ACC    &BAC    &AUC   &SEN   &SPE  &$p<0.05$\\
\midrule
Baseline    &64.12      &64.04      &69.20      &63.47      &64.60  &Yes\\
TCA         &60.56      &60.71      &62.24      &61.68      &59.73  &Yes\\   
GFK         &65.14      &64.92      &70.08      &63.47     &66.37  &Yes\\
SA          &65.14      &64.92      &70.09      &63.47      &66.37  &Yes\\
CORAL       &64.38      &64.26      &69.84      &63.47      &65.04  &Yes\\
VoxCNN      &70.51      &69.89      &74.74  &\textbf{80.61}   &59.18  &Yes\\
IADT (Ours)       &\textbf{73.54}   &\textbf{73.00} &\textbf{75.70}  &69.46   &\textbf{76.55}  &--\\
\bottomrule[1.2pt]
\end{tabular}
}
\label{tab:MCI}
\end{table}
\section{Discussion} \label{Discussion}

\subsection{Data Distribution Visualization} 
\begin{figure}[t]
\center
 \includegraphics[width= 0.49\linewidth]{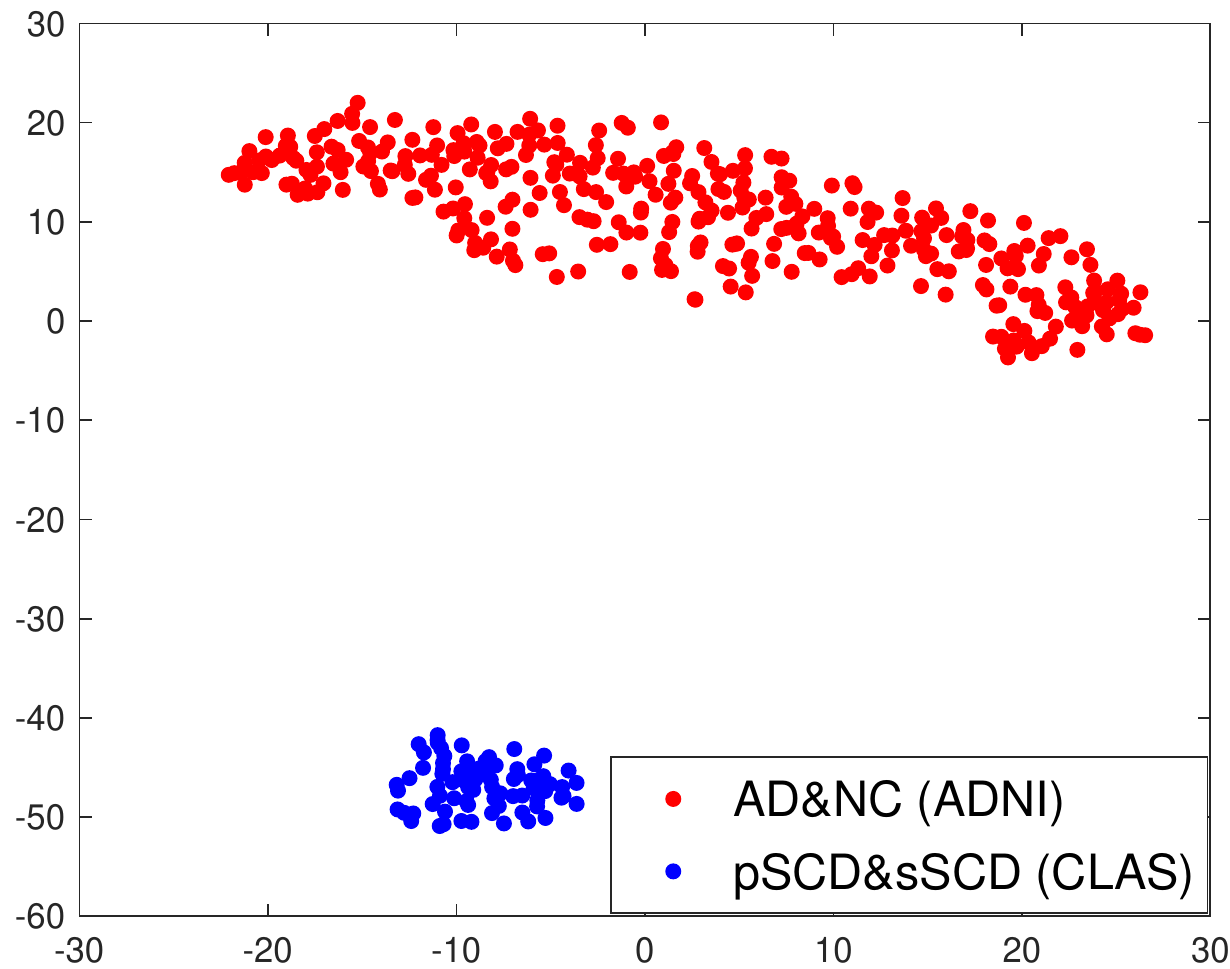}
 \includegraphics[width= 0.49\linewidth]{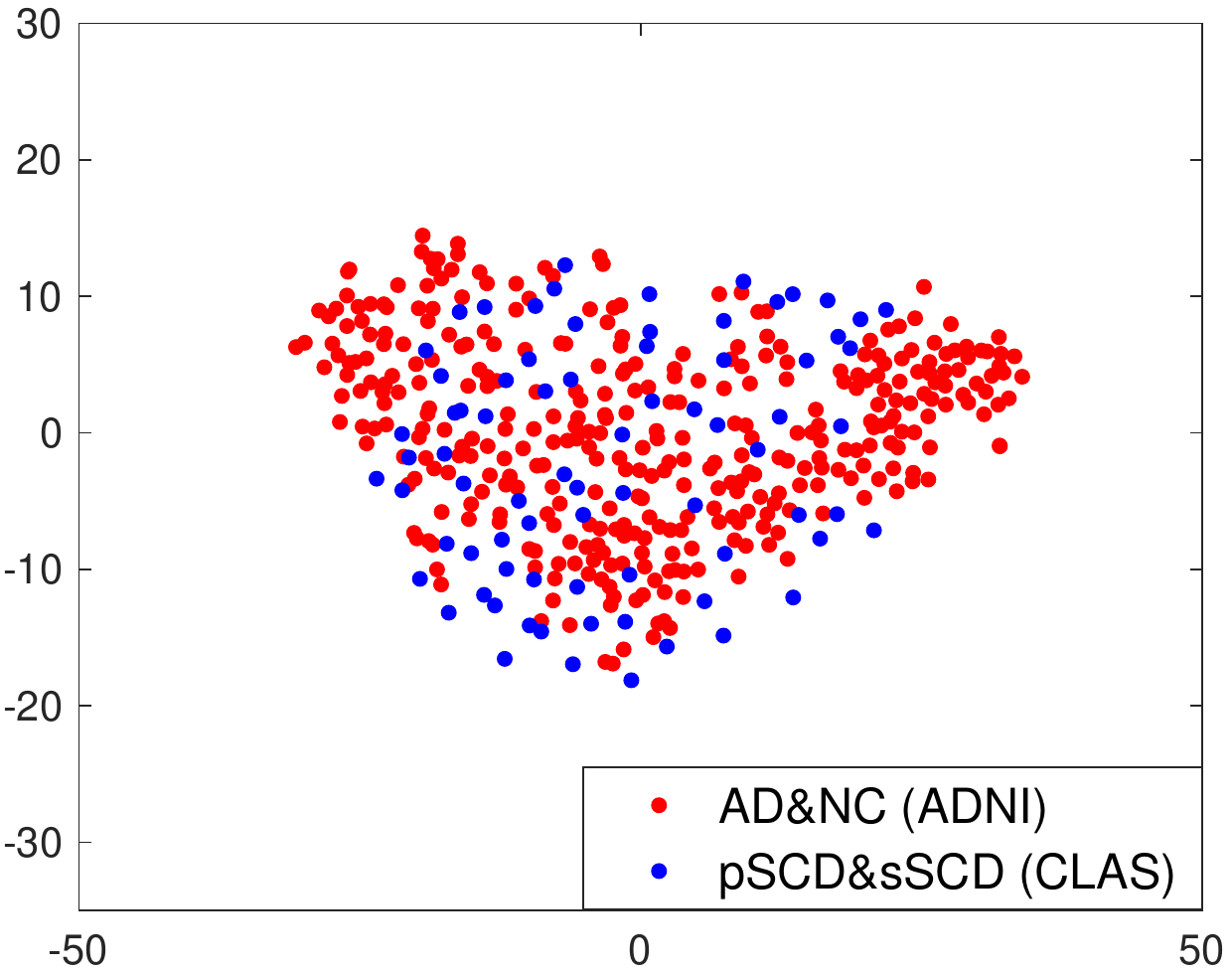}
 \caption{ Data distribution of the source (AD\&NC samples from ADNI) and target data (SCD samples from CLAS) before and after the domain adaptation processing.}
 \label{fig_distribution}
\end{figure}

We extract MRI features of source data (AD\&NC samples from ADNI) and target data (SCD samples from CLAS) learned by the encoder. 
The source and target data (represented by ROI-based MRI features) distribution before and after domain adaptation is visualized using t-SNE~\cite{tsne}, as shown in Fig.~\ref{fig_distribution}.
From Fig.~\ref{fig_distribution}, the two domains have significant distribution differences in the original space. 
After adaptation, the proposed encoder is able to learn some shared features that can decrease the domain shift between two datasets. 

\subsection{Discriminative Brain Regions}
\begin{figure}[t]
\setlength{\belowcaptionskip}{1pt}
\setlength{\abovecaptionskip}{1pt}
\center
 \includegraphics[width= 0.96\linewidth]{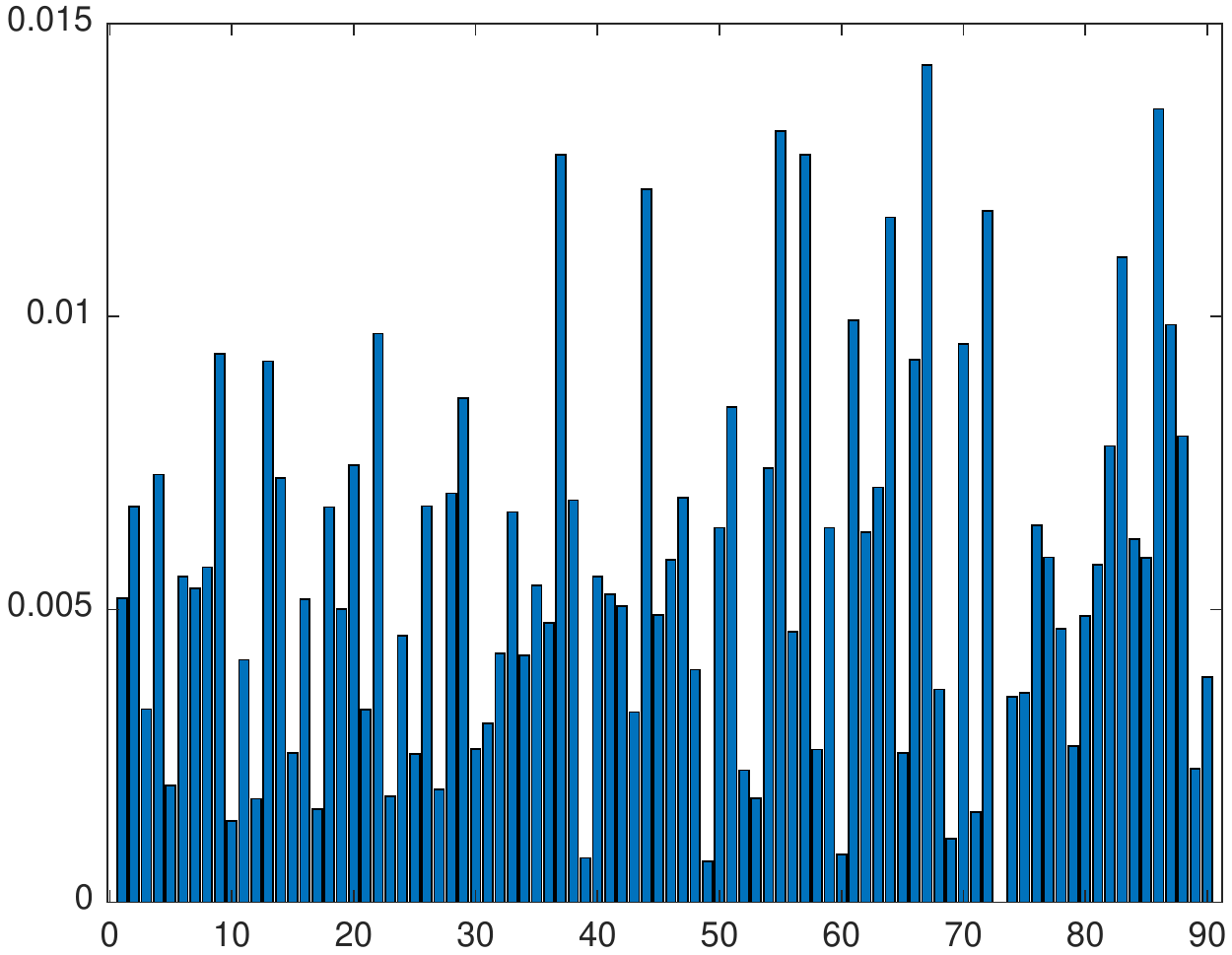}
 \caption{Discriminative power (weights) of different brain ROIs outputted by the attention module of our model for progressive SCD identification.}
 \label{fig_ROIs}
\end{figure}

It is helpful to identify the brain regions that are more closely linked to SCD progression. 
Thus we explore the top ten brain ROIs automatically identified by the attention module of our model. 
We first aggregate the progressive SCD instances that have been correctly identified by our model.
Then they are fed into the network and get their attention vectors which are computed by the attention module. 
Each element in the attention vector reflects the corresponding feature's importance.
The mean value of the attention vectors is used to indicate the brain region contribution (for visualization, we also minus the minimum value of the vector for each element).
The discriminative ROIs selected by our model for pSCD identification on CLAS and pMCI vs. sMCI classification on ADNI are shown in Fig.~\ref{fig_ROIs} and Fig.~\ref{fig_ROIs_MCI}, respectively.
We list the top ten ROIs for pSCD identification (first two columns) and pMCI vs. sMCI classification (last two columns) in Table~\ref{tab:ROIs}.
We also visualize five discriminative brain regions (based on AAL) located by our method for pSCD identification on CLAS and MCI-to-AD conversion prediction on ADNI in Fig.~\ref{fig_brain_regions}.

From Table~\ref{tab:ROIs}, it can be observed that the most informative brain regions include the precuneus, hippocampus and certain gyrus areas. 
Especially, the precuneus is a brain region involved in a variety of cognitive functions, which include episodic memory retrieval, mental imagery strategies, and consciousness~\cite{lundstrom2003isolating,cavanna2006precuneus,hebscher2019causal}.
It is recently reported as a critical brain area for the memory and cognition impairment observed in the early stage of AD~\cite{karas2007precuneus,scheff2013synapse,koch2018transcranial}. 
Also, the hippocampus is a complex brain structure which lies in the temporal lobe. It plays a  major role in memory and learning~\cite{eichenbaum1992hippocampus,hannula2012hippocampus,gonzalez2019persistence}. Studies have revealed that it can be affected in a variety of neurological disorders such as Alzheimer's disease~\cite{hyman1984alzheimer,shi2009hippocampal,perrotin2015hippocampal}. 
In addition, we also find an interesting phenomenon from our results that progressive SCD is more closely linked with the left brain areas rather than the right part. This has also been found by some related studies~\cite{yue2021prediction}.

\begin{table}[t]
\setlength{\belowcaptionskip}{1pt}
\setlength{\abovecaptionskip}{1pt}
\caption{Ten discriminative brain ROIs selected by the attention module of our model for progressive SCD identification (\ie, pSCD vs. sSCD classification) on CLAS and MCI-to-AD conversion prediction (\ie, pMCI vs. sMCI classification) on ADNI.}
\centering
\setlength{\tabcolsep}{2pt}{
\begin{tabular} {cc c|c cc} 
\toprule[1.2pt]
\multicolumn{2}{c}{pSCD vs. sSCD}  && &  \multicolumn{2}{c}{pMCI vs. sMCI}\\
\cline{1-2} 
\cline{5-6}
Index   &ROI Name  &&   &Index    &ROI Name  \\

\midrule

67      &Precuneus left       &&            &32  &Anterior cingulate     \\
86      &Middle temporal gyrus right   &&   &35  &Posterior cingulate gyrus left \\
55      &Fusiform gyrus  left          &&   &8   &Middle frontal gyrus right  \\
37      &Hippocampus left              &&   &90  &Inferior temporal gyrus right\\
57      &Postcentral gyrus left        &&   &6   &Superior frontal gyrus \\
44      &Calcarine fissure right       &&   &37  &Hippocampus left\\
72      &Caudate nucleus right         &&   &76  &Lenticular nucleus\\
64      &Supramarginal gyrus right     &&   &21  &Olfactory cortex left\\
83      &Temporal pole left            &&   &49  &Superior occipital gyrus left\\
61      &Inferior parietal left        &&   &66  &Angular gyrus right\\
\bottomrule[1.2pt]
\end{tabular}
}
\label{tab:ROIs}
\end{table}

\begin{figure}[t]
\setlength{\belowcaptionskip}{1pt}
\setlength{\abovecaptionskip}{1pt}
\center
 \includegraphics[width= 0.9\linewidth]{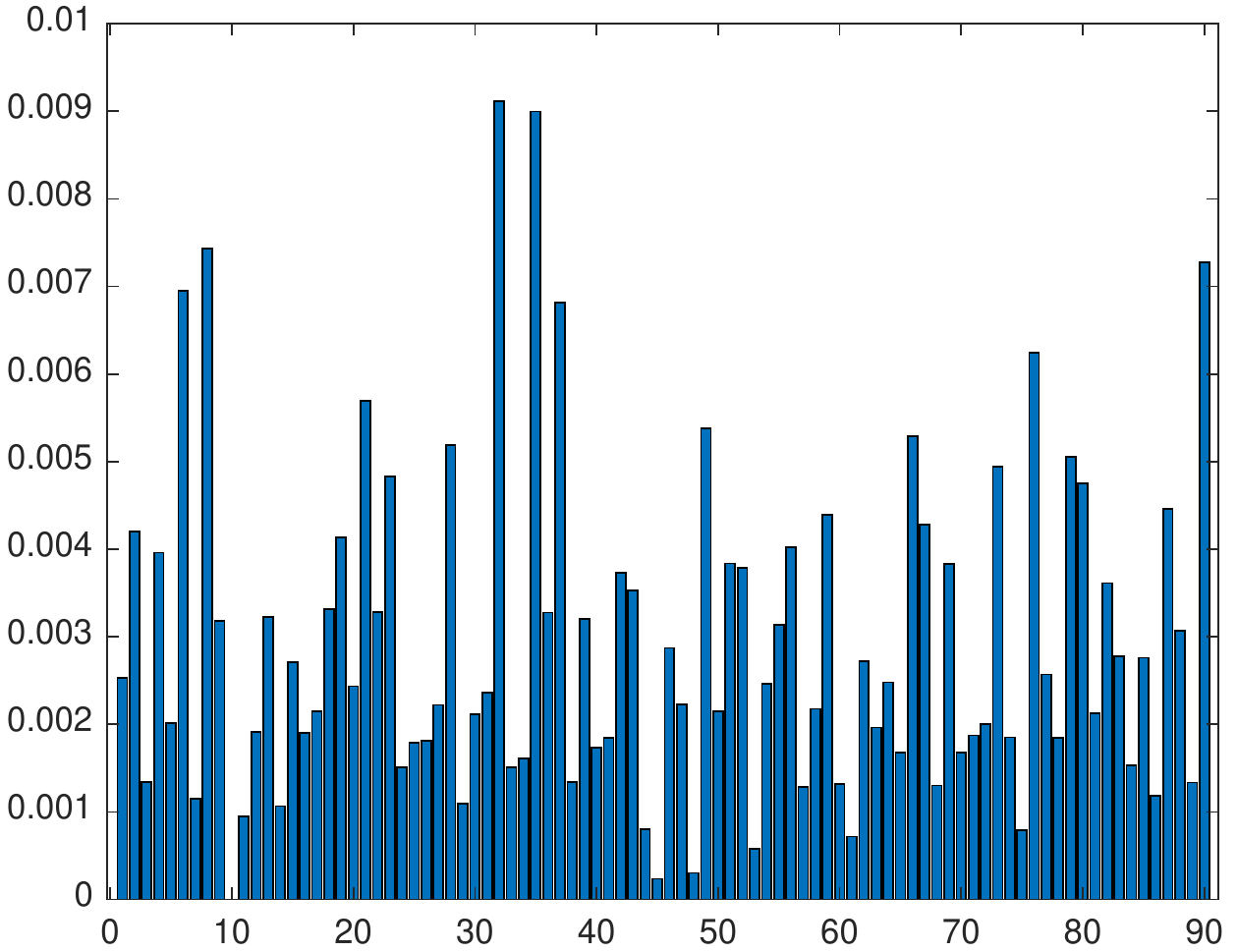}
 \caption{Discriminative power (weights) of different Brain ROIs outputted by the attention module of our model for MCI-to-AD conversion prediction (\ie, pMCI vs. sMCI classification).}
 \label{fig_ROIs_MCI}
\end{figure}

\if false
\begin{figure}[t]
\center
 \includegraphics[width= 1.0\linewidth]{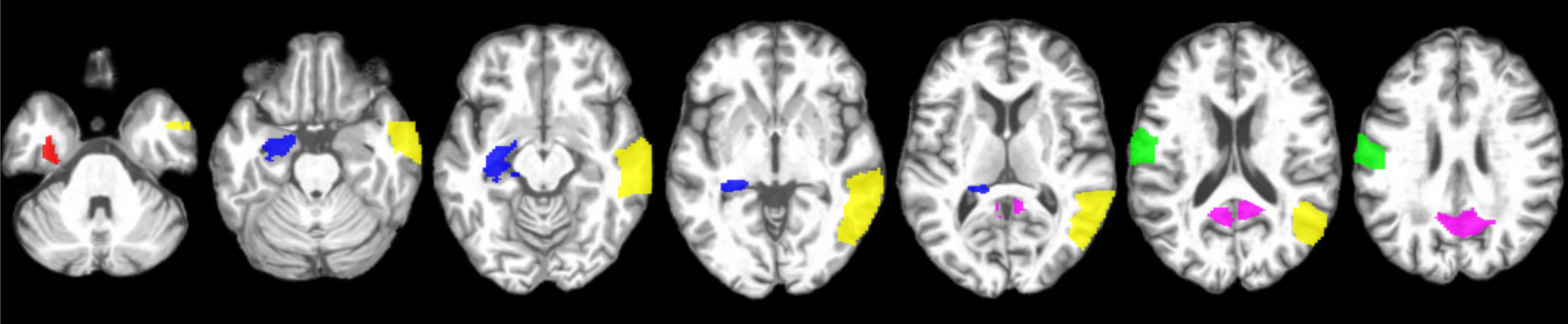}
 \includegraphics[width= 1.0\linewidth]{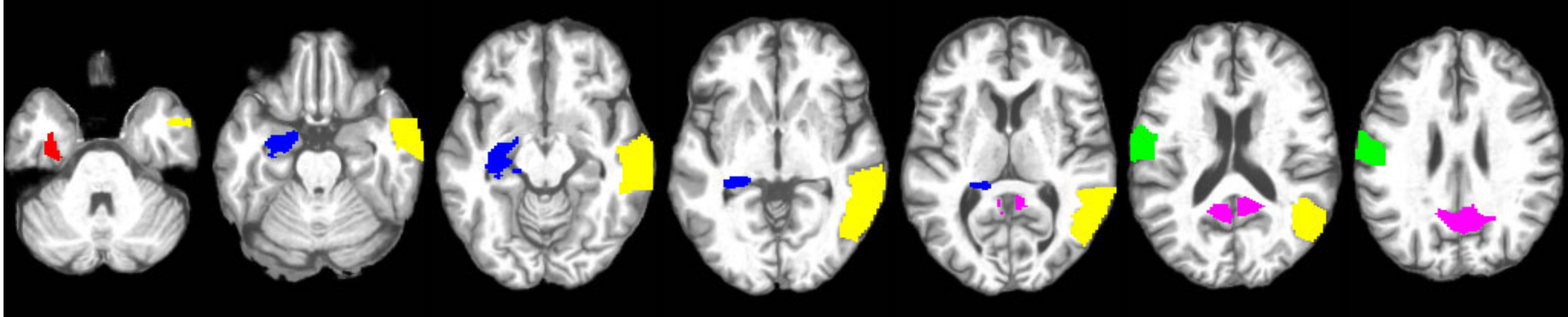}
 \caption{Import brain regions identified by the proposed IADT method in pSCD vs. sSCD classification in CLAS.}
 \label{fig_brain_regions}
\end{figure}
\fi

\begin{figure}[t]
\setlength{\belowcaptionskip}{1pt}
\setlength{\abovecaptionskip}{1pt}
\centering
\includegraphics[width= 1.0\linewidth]{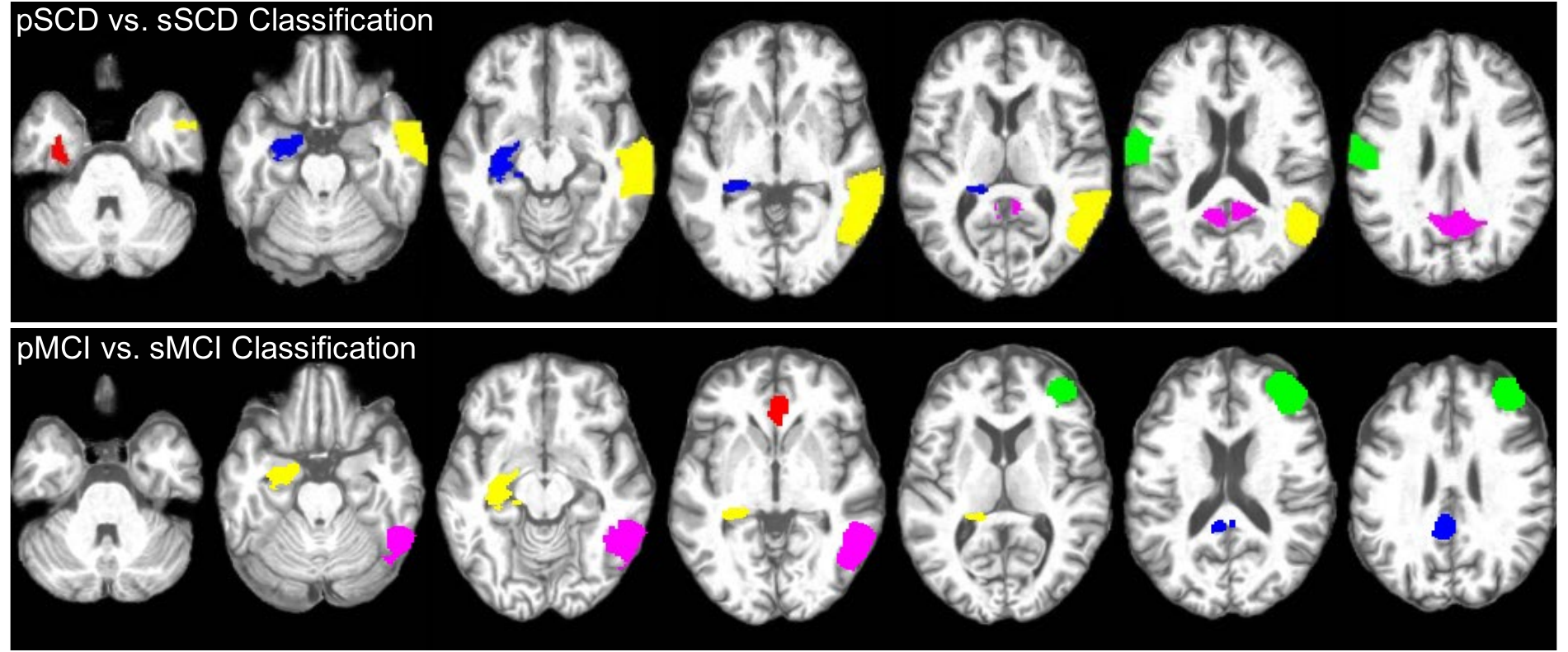}
\caption{Important brain regions identified by the proposed IADT in progressive SCD identification on CLAS (top) and MCI-to-AD conversion prediction on ADNI (bottom).}
\label{fig_brain_regions}
\end{figure}

\subsection{Ablation Study}
\begin{figure}[t]
\setlength{\abovecaptionskip}{1pt} 
\centering
\begin{subfigure}{.45\textwidth}
    \centering
    \includegraphics[width=0.8\textwidth]{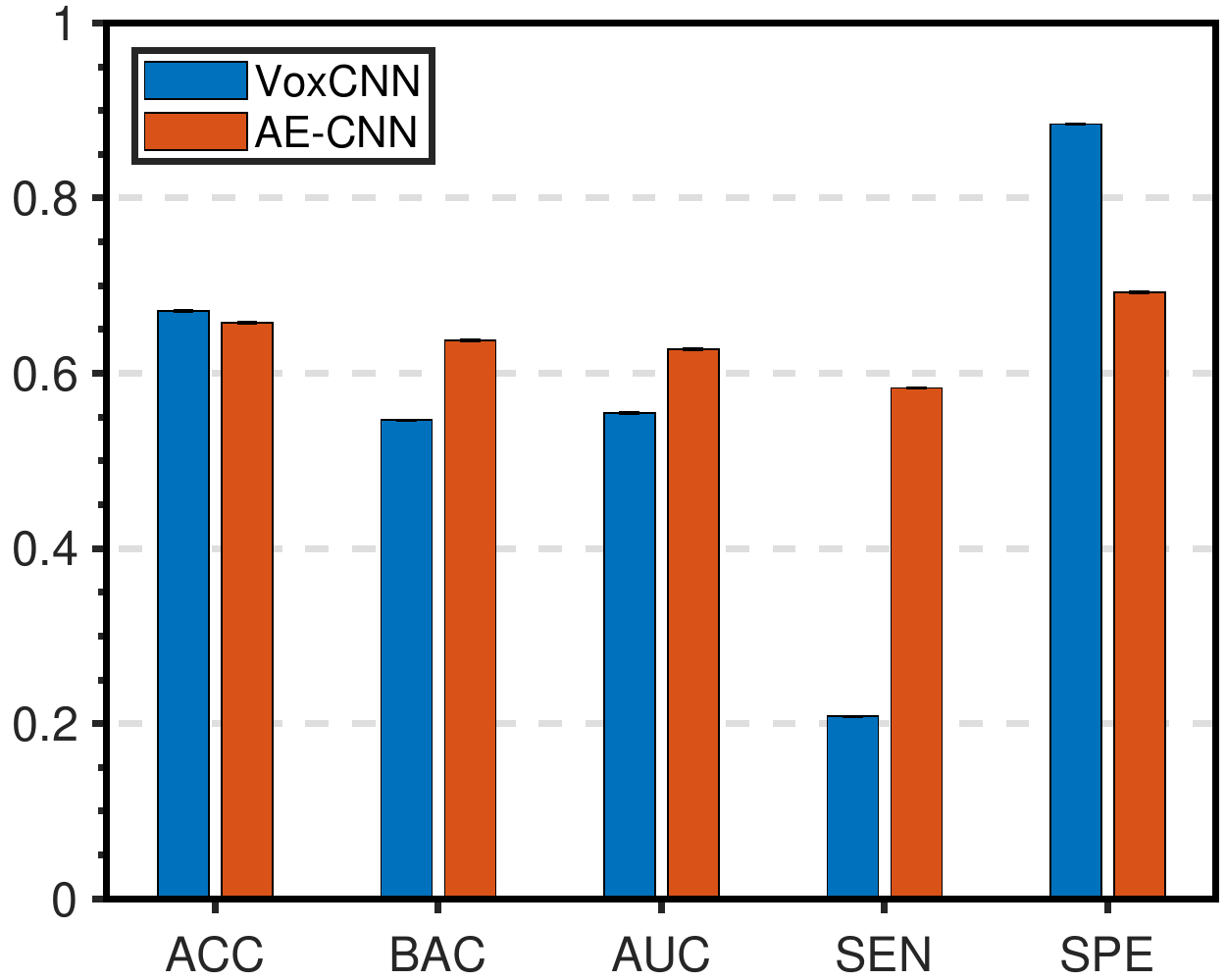}
    \caption{Performance comparison of VoxCNN and our model with CNN features in the task of SCD progression prediction.}
    \label{fig_CNN1}
\end{subfigure}
\begin{subfigure}{.45\textwidth}
    \centering
    \includegraphics[width=0.8\textwidth]{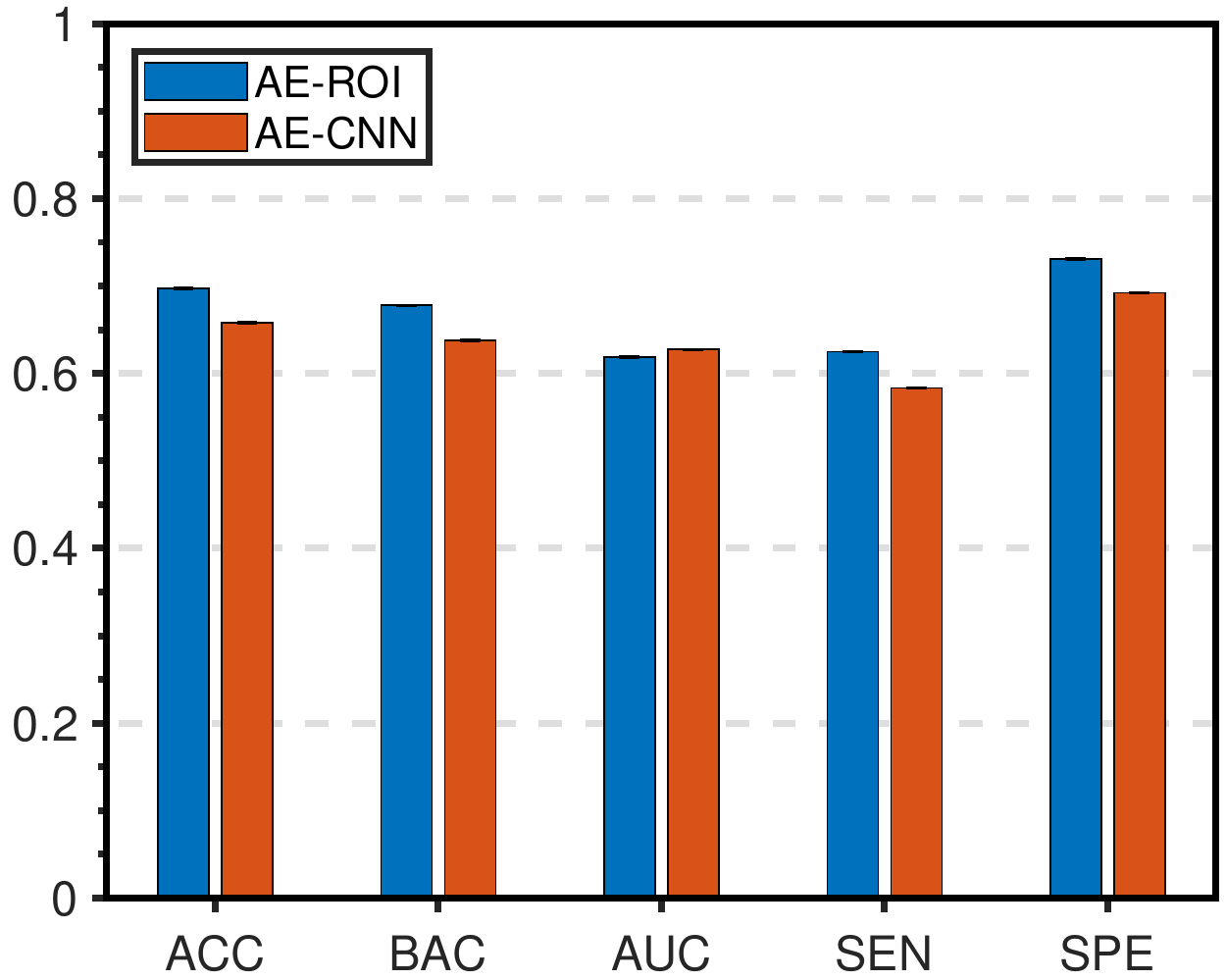}
    \caption{Performance comparison of the proposed model with ROI features and CNN features for SCD progression prediction.}
\label{fig_CNN2}
\end{subfigure}
   \caption{Ablation study of our method with CNN features.}
\label{fig_CNN}
\end{figure}

To investigate the classification performance with other MRI features, we further use CNN features as the input of the proposed IADT model, and test its performance for pSCD vs. sSCD classification.
Specifically, we use a 3D CNN which is pretrained with brain MRIs from other data sources as the feature extractor. 
Then the 3D CNN extracts features of the source data (AD/NC samples) and target data (SCD samples), respectively. 
The learned features are then fed into our model for adaptation and classification.
In practice, we adopt a widely-used 3D VGG-Net CNN for brain MRI classification network, \ie, VoxCNN~\cite{VoxCNN}. 
It is pretrained with 205 AD and 231 NC samples acquired from other independent subjects in ADNI. 
The network involves $10$ convolution layers, and the embeddings from the last activation of the convolution layer are used as the CNN feature. 
The dimension of CNN feature is $256$.
We also apply VoxCNN to pSCD vs. sSCD classification, and compare the result with our method, as shown in Fig.~\ref{fig_CNN1}.
It can be observed that the proposed model outperforms the CNN in most cases. 
The CNN model has a trend of overfitting to the negative samples (sSCD) whereas our model is able to achieve a much more balanced accuracy. 
We also compare our method with ROI features and CNN features, as shown in Fig.~\ref{fig_CNN2}.
From the results, the model with CNN features achieves higher AUC values while ROI features enable the model to achieve higher classification accuracy. 
Overall, their performances are comparable without dominant advantages over each other in SCD progression prediction.

\subsection{Parameter Analysis}
\subsubsection{Influence of Dimension of Latent Space}
\begin{figure}[!tbp]
\setlength{\belowcaptionskip}{1pt}
\setlength{\abovecaptionskip}{1pt}
\center
 \includegraphics[width= 0.8\linewidth]{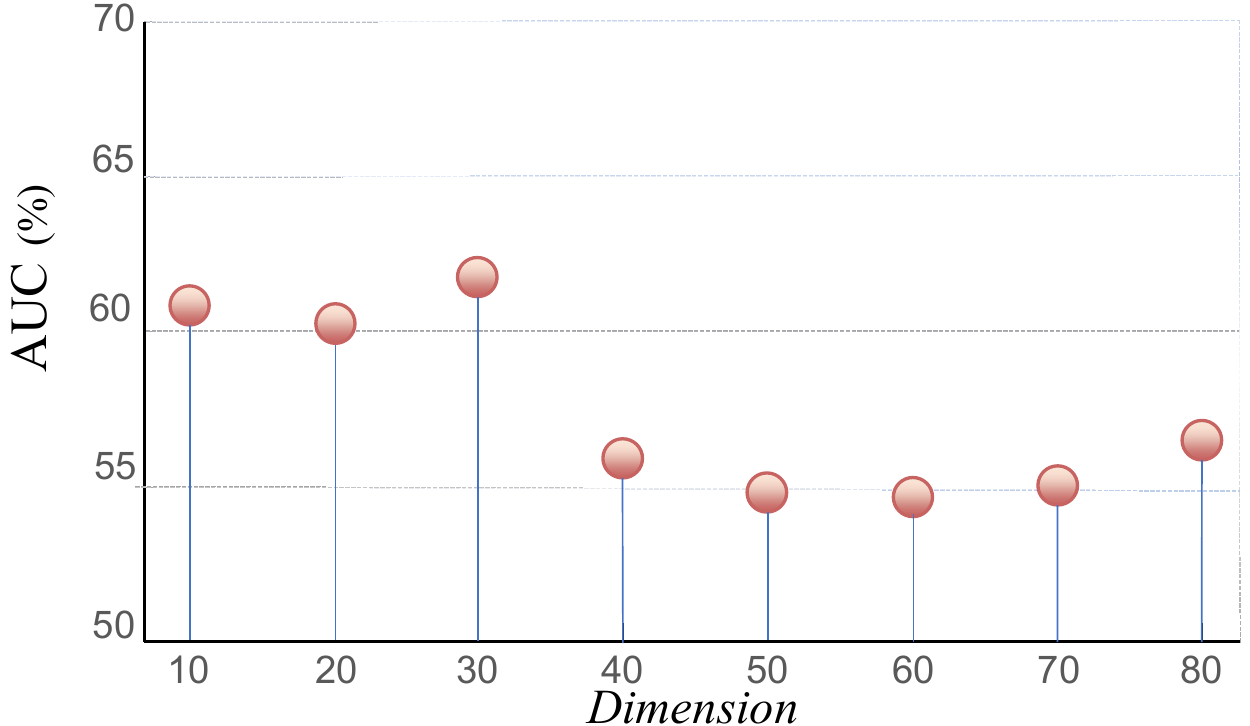}
 \caption{Influence of dimension of the latent feature space on pSCD vs. sSCD classification.}
 \label{fig_Dimension}
\end{figure}

The encoder in our model plays the role of projecting source and target data into a shared latent space. Since this space is a compressed representation of the original features, the dimension of latent space is tunable. 
We explore the influence of this dimension and calculate the AUC values under different dimensions. The result is reported in Fig.~\ref{fig_Dimension}.
It can be observed that the model can achieve relatively good classification performance when the dimension of latent space is between 10$\thicksim$40. 
This implies that low-dimension spaces can be explored for cross-domain adaptation in the task of SCD progression prediction.

\subsubsection{Influence of Losses}
\begin{figure}[!tbp]
\setlength{\belowcaptionskip}{1pt}
\setlength{\abovecaptionskip}{1pt}
\center
 \includegraphics[width= 0.9\linewidth]{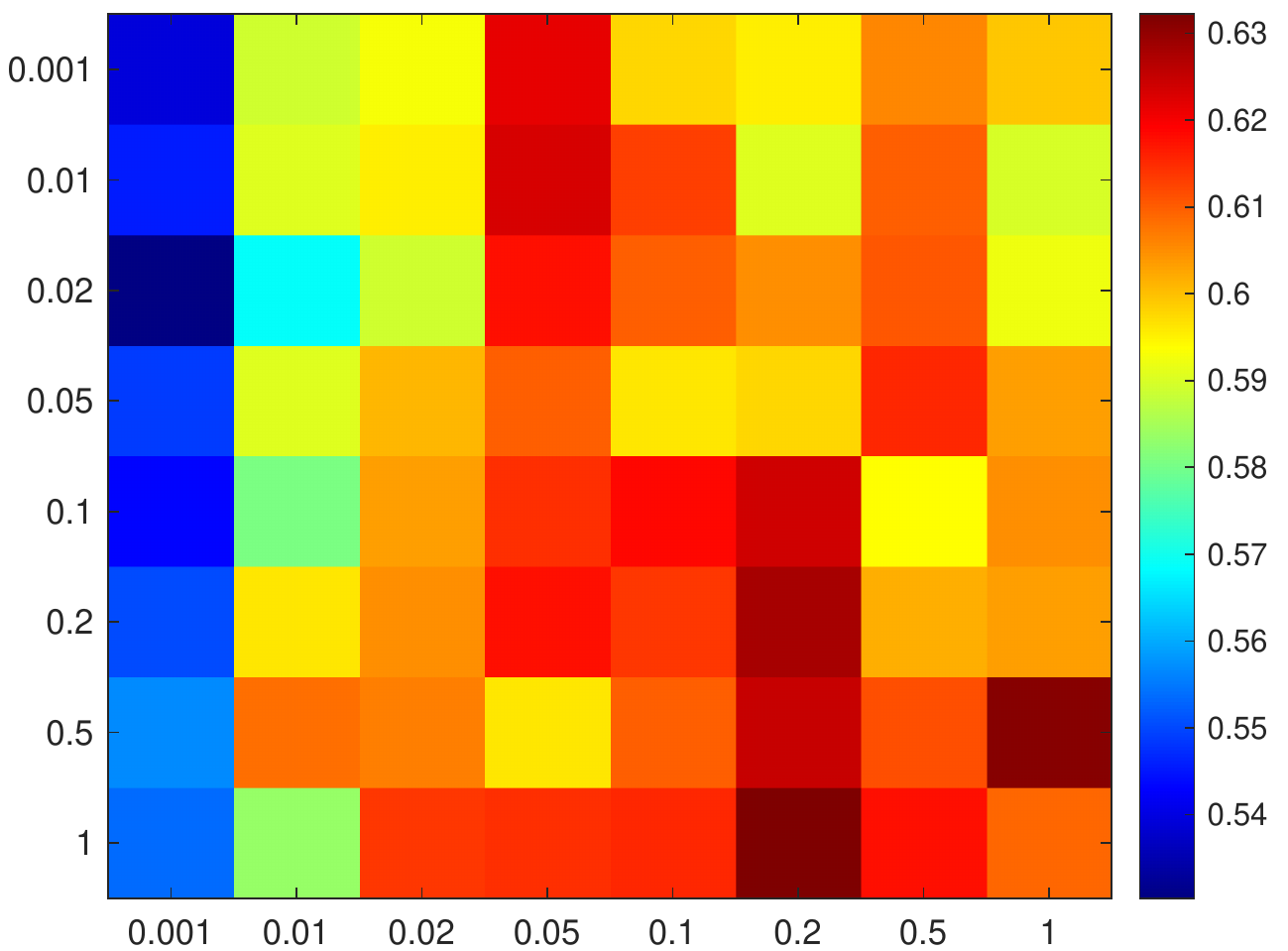}
 \caption{Influence of the parameter of $\lambda_1$ (y-axis) and $\lambda_2$ (x- axis) on pSCD vs. sSCD classification in terms of AUC values.}
 \label{fig_AUC_parameter}
\end{figure}

There are three losses for our model training as shown in Eq.~\ref{overall_loss}. 
Since the target data (SCD) number is tiny, the reconstruction loss $\mathcal{L}_{recon}$ plays an important role in reflecting the target data information.
Thus we fix this parameter to $1$ in Eq.~\ref{overall_loss} for the reconstruction loss, and vary the values of $\lambda_1$ and $\lambda_2$ to explore their influences.
Specifically, we independently change the values of $\lambda_1$ and $\lambda_2$ from {0.001, 0.01, 0.02, 0.05, 0.1, 0.2, 0.5, 1}, and compute the corresponding AUC values for the pSCD vs. sSCD classification task.
The result is reported in Fig.~\ref{fig_AUC_parameter}.

From Fig.~\ref{fig_AUC_parameter}, we can see that when the $\lambda_2$ is relatively small ($\leq$0.02), the overall classification performance in terms of AUC is not good. 
This indicates the importance of the classification module in our model. 
When its contribution is reduced, the model will lose discrimination toward brain disorders.
In addition, in most cases, the classification performance with respect to  $\lambda_1$ and  $\lambda_2$ is stable, 
which demonstrates that our model is not very sensitive to parameters.

\section{Conclusion} \label{Conclusion}
In this paper, we propose an interpretable autoencoder model with domain transfer (IADT) for SCD progression prediction. 
Our model takes brain ROI features as the input. An attention module helps automatically find the most discriminative brain disorder-related regions. The domain adaptation and the joint training of the classification and reconstruction modules help the model learn domain-invariant features. 
Our model has three advantages: 1) good performance on small-sample-sized datasets which has extensive applications in medical imaging; 
2) good interpretability which can help analyze disease-related brain areas with different brain atlases; 
and 3) simple architecture and fast training/running speed. 
It only takes 5$\thicksim$10 seconds on CPUs to train and test the model with reproducible results. In the future, we will incorporate multi-modality features into our framework for further analysis.

\section*{Acknowledgment}
H.~Guan, P.-T.~Yap, A.~Bozoki and M.~Liu were supported in part by NIH grant RF1AG073297. 
Part of the data used in this paper were obtained from the Alzheimer's Disease Neuroimaging Initiative (ADNI) database and from the Chinese Longitudinal Aging Study (CLAS) database. 
The investigators within the
ADNI contributed to the design and implementation of ADNI
and provided data but did not participate in analysis or writing of this article. 
\ifCLASSOPTIONcaptionsoff
  \newpage
\fi

\footnotesize
\bibliographystyle{IEEEtran}
\bibliography{mybib}

\end{document}